\documentclass{egpubl}
\usepackage{pg2026}

\usepackage[T1]{fontenc}
\usepackage{dfadobe}

\usepackage{cite}
\BibtexOrBiblatex

\electronicVersion
\PrintedOrElectronic

\ifpdf
  \usepackage[pdftex]{graphicx}
\else
  \usepackage[dvips]{graphicx}
\fi

\usepackage{egweblnk}

\usepackage{amsmath}
\usepackage{amssymb}
\usepackage{booktabs}
\usepackage{multirow}

\title[MGAvatar]%
      {MGAvatar: Mesh-Bound Gaussians for Head Avatar Geometry and Appearance Modeling}

\author[L. Shi et al.] {\parbox{\textwidth}{\centering  Lei Shi$^{1}$\orcid{0009-0008-1095-1295}, Sen Peng$^{1}$\orcid{0009-0001-3886-2998}, Zhiyang Deng$^{1}$\orcid{0009-0005-5105-1400}, Zhonggui Chen$^{2}$\orcid{0000-0002-9960-4896}, Xiaohu Guo$^{3}$\orcid{0000-0002-2141-7222}, Baorong Yang$^{4}$\orcid{0000-0002-2896-2506}, and Xiao Dong$^{1}$\thanks{Corresponding author}\orcid{0000-0003-4365-0575} } \\ {\parbox{\textwidth}{\centering 
$^1$Guangdong Provincial/Zhuhai Key Laboratory of IRADS, Beijing Normal-Hong Kong Baptist University, China\\ $^2$School of Informatics, Xiamen University,  China\\ $^3$Department of Computer Science, The University of Texas at Dallas, USA\\ $^4$College of Computer Engineering, Jimei University, China } } } 

\begin{document}


\maketitle
\begin{abstract}
   Accurate head modeling requires a stable yet expressive geometric representation. Existing Gaussian-based head avatars commonly rely on parametric templates (e.g., FLAME) for Gaussian initialization and deformation, but these templates lack personalized priors and struggle to represent structures such as hair and clothing. To address this issue, we propose MGAvatar, a Gaussian–mesh hybrid representation that jointly models geometry and appearance through two Gaussian–mesh binding modes. Specifically, we introduce vertex-bound Gaussians and constrain their learnable parameters, enabling progressive mesh deformation to represent complex head geometry, while a pose-dependent offset module accounts for non-rigid deformations. Once geometry is stabilized, MGAvatar switches to face-bound Gaussians for appearance modeling. To improve appearance consistency across novel poses and viewpoints, we introduce a view-conditioned neural color field that alleviates artifacts caused by independently optimized Gaussian colors. In addition, we design a Gaussian offset network to predict Gaussian offset maps in the observation space, providing greater flexibility for face-bound Gaussians to capture dynamic facial textures. Extensive experiments on multi-view and monocular videos show that MGAvatar outperforms existing methods in rendering quality, producing high-fidelity head avatars with rich texture details. Our project page is available at \textcolor{blue}{https://bnbucv.github.io/MGAvatar/}.


\end{abstract}  

\section{Introduction}
High-fidelity and animatable head avatar reconstruction has long been a central topic in computer vision and computer graphics, with wide-ranging applications in AR/VR~\cite{he2020collabovr}, digital humans~\cite{demirel2007applications}, and gaming~\cite{waggoner2009my}.
These methods aim to recover detailed 3D geometry and appearance from images or videos, while enabling realistic animation under varying expressions, poses, and lighting. 
Recent advances in 3D reconstruction have explored a variety of scene representations for both static and dynamic modeling. Implicit representations, such as neural radiance fields (NeRF)~\cite{mildenhall2021nerf}, encode scenes as continuous neural fields and demonstrate impressive performance in novel view synthesis. However, their reliance on volumetric rendering typically leads to high computational and memory costs. More recently, 3D Gaussian Splatting (3DGS)~\cite{kerbl20233d} proposes to use discrete 3D Gaussians with differentiable rasterization, enabling significantly faster rendering while preserving high-fidelity visual details. Building upon these advances, researchers have developed a variety of head avatar reconstruction approaches, including mesh-based~\cite{lombardi2018deep,ma2021pixel,grassal2022neural}, NeRF-based~\cite{gafni2021dynamic,lombardi2021mixture,gao2022reconstructing}, point-based~\cite{wang2023neural,zheng2023pointavatar}, and 3D Gaussian-based~\cite{dhamo2024headgas,qian2024gaussianavatars} approaches, each exploring different trade-offs between efficiency, fidelity, and animatability.

The human head exhibits complex geometry and non-rigid motion, making accurate reconstruction challenging. To address this, many 3DGS-based methods~\cite{qian2024gaussianavatars,shao2024splattingavatar} leverage 3D Morphable Models (3DMMs)~\cite{li2017learning,loper2023smpl,pavlakos2019expressive} as geometric priors to capture facial expressions and head motion. Although these methods achieve high-quality reconstruction, they still struggle with complex regions such as hair and shoulders, because the FLAME template does not model beyond-template regions well, as well as non-rigid deformations of the face.
GaussianAvatars~\cite{qian2024gaussianavatars} proposes a VHAP~\cite{qian2024vhap} head tracking preprocessing pipeline to obtain coarse geometric offsets. However, these offsets are restricted to the head region and cannot model structures such as the shoulders, and under single-view settings, the estimated geometry is unreliable.
Some methods~\cite{wang2025mega,liao2025hhavatar} propose to decompose head reconstruction into separate components, estimating hair Gaussians, which are then fused with FLAME-based head Gaussians. These approaches require complex preprocessing  and cannot leverage the FLAME prior to drive hair motion, leading to increased computational cost and  inaccuracies. Moreover, existing methods often suffer from  unstable appearance under novel views and poses, which may lead to noticeable artifacts.

To address the limitations of FLAME-based 3DGS head reconstruction—such as neutral templates lacking personalized geometry, hair rendering artifacts, and view-dependent color instability—we propose MGAvatar, a hybrid mesh-bound Gaussian representation that jointly models geometry and appearance for animatable, high-fidelity head avatars. Specifically, to capture structures beyond the FLAME template, we bind Gaussians to the mesh and introduce a reduced degree-of-freedom (DoF) design for Gaussian attributes to enable flexible mesh adjustment. This design allows the model to capture both static geometric features and non-rigid head deformations, providing stronger geometric priors for subsequent rendering. To improve color consistency across novel viewpoints and poses, we replace the SH-based color representation with a neural color field, enhancing appearance stability and reducing rendering artifacts. The mesh optimized through geometric learning can be directly driven by FLAME parameters, enabling full-head animation—including hair and shoulders—while maintaining the benefits of a unified representation. In summary, the main contributions of our method are as follows:
\begin{itemize}
\item We propose a hybrid avatar representation that combines vertex-bound and face-bound Gaussians to jointly model geometry and appearance for animatable head avatars.

\item We develop a reduced-DoF geometric learning strategy for vertex-bound Gaussians, enabling flexible mesh refinement to capture identity-specific geometry.

\item We introduce a view-conditioned neural color field for face-bound Gaussians, together with Gaussian attribute offset maps, to model fine-grained appearance details and improve appearance consistency across novel viewpoints.

\item We validate our method on both multi-view and monocular datasets, demonstrating significant improvements over state-of-the-art methods in novel-view synthesis and avatar reenactment.
\end{itemize}

\section{Related Work}

\textbf{Head avatar representations.} 
Head avatar modeling has evolved toward more realistic and flexible representations. Existing head avatar methods primarily relied on 3D Morphable Models (3DMMs)~\cite{blanz2023morphable,gerig2018morphable}, which provide compact parametric representations of identity, pose, and expression.
However, parametric templates are limited to modeling fine facial details and hair regions. To address these limitations, several works~\cite{cao2013facewarehouse, li2017learning, tran2018nonlinear, ploumpis2020towards} extend 3DMMs to improve the representation ability. Some works~\cite{chaudhuri2020personalized, bai2021riggable,khakhulin2022realistic, bharadwaj2023flare,retsinas20243d} exploit explicit mesh to model dynamic head avatars. NHA~\cite{grassal2022neural} designs two feed-forward networks to optimize geometry and predict colors. FLARE~\cite{bharadwaj2023flare} combines mesh-based differentiable rendering with neural networks for deformation, materials, and illumination modeling.

With the advent of NeRF~\cite{mildenhall2021nerf}, a series of methods~\cite{athar2023flame, bai2023learning, xu2023latentavatar,duan2023bakedavatar, zielonka2023instant} explore volumetric representations for head avatars. IMAvatar~\cite{zheng2022avatar} and HeadNeRF~\cite{hong2022headnerf} leverage implicit neural representations for head reconstruction and appearance modeling, while NeRFBlendShape~\cite{gao2022reconstructing} represents a head as a weighted combination of radiance field bases according to expression coefficients.
NeuFace~\cite{zheng2023neuface} and SFDM~\cite{jin2025sfdm} investigate realistic face rendering by modeling facial reflectance and disentangling geometry from reflectance.
PointAvatar~\cite{zheng2023pointavatar} represents avatars as deformable 3D point-based models, offering increased topological flexibility and expressive capacity compared with NeRF-based methods. DELTA~\cite{feng2023learning} proposes a hybrid model to leverage mesh and NeRF to model faces and hair separately.

Following the introduction of 3DGS~\cite{kerbl20233d}, many studies~\cite{peng2025genericavatar, xiang2024flashavatar, peng2025rmavatar} focus on digital human reconstruction based on Gaussian representations. MonoGaussianAvatar~\cite{chen2024monogaussianavatar} models deformable head avatars with learnable 3D Gaussians.
Hybrid Explicit Representation~\cite{cai2024hybrid} combines UV-mapped meshes with 3D Gaussians, while SVG-Head~\cite{sun2025svg} introduces surface and volumetric Gaussians bound to meshes for head reconstruction.
GaussianHead~\cite{wang2025gaussianhead} adopts tri-plane to store appearance related attributes. GaussianBlendShape~\cite{ma20243d} and RGBAvatar~\cite{li2025rgbavatar} combine a neutral model with expression blendshapes via linear blending of Gaussians to generate heads with arbitrary expressions. 
MeGA~\cite{wang2025mega} and HHAvatar~\cite{liao2025hhavatar} represent the head and hair separately using 3D Gaussians, but require additional preprocessing and hair motion estimation.
Fate~\cite{zhang2025fate}, STAvatar~\cite{zhao2026stavatar} and UIKA~\cite{wu2026uika} further exploit UV-space features to predict Gaussian attributes, enabling high-quality avatar reconstruction.

\textbf{3D Gaussians with 3DMMs.} Recent works have increasingly integrated 3DMMs with Gaussian splatting to exploit geometry priors for animatable head reconstruction. Prior studies adopt diverse strategies to inject 3DMM guidance, including template-based initialization, UV-guided sampling, and Gaussian-mesh binding. For example, GaussianAvatar~\cite{hu2024gaussianavatar} initializes Gaussian primitives from sampled points on the SMPL template. FlashAvatar~\cite{xiang2024flashavatar} conducts UV sampling to locate Gaussian positions on the mesh surface. GaussianAvatars~\cite{qian2024gaussianavatars}, GoMAvatar~\cite{wen2024gomavatar} and SplattingAvatar~\cite{shao2024splattingavatar} propose hybrid representations which bind Gaussians to mesh faces so that they move consistently with the underlying mesh while still allowing limited Gaussian attribute adaptation.

Several works further optimize 3DMM parameters to better guide Gaussian representations. For example, ExpressiveAvatar~\cite{moon2024expressive} and GaussianAvatars~\cite{qian2024gaussianavatars} optimize pose parameters. 
GaussianHeadAvatar~\cite{xu2024gaussian} estimates 3DMM parameters and predicts Gaussian displacements from pose and expression. 
Inspired by NHA~\cite{grassal2022neural}, 
MeGA~\cite{wang2025mega} learns UV-space displacements via a CNN to refine the mesh; however, the inherent smoothness of the network limits the degree of mesh deformation, making it difficult to model complex hair geometry. 
Unlike previous methods relying on fixed template priors, we enhance FLAME mesh expressiveness through template adaptation and deformation modeling, while tightly coupling Gaussians with the mesh to leverage optimized geometry and Gaussian rendering flexibility.
\begin{figure*}[htb]
  \centering
  \includegraphics[width=1\textwidth]{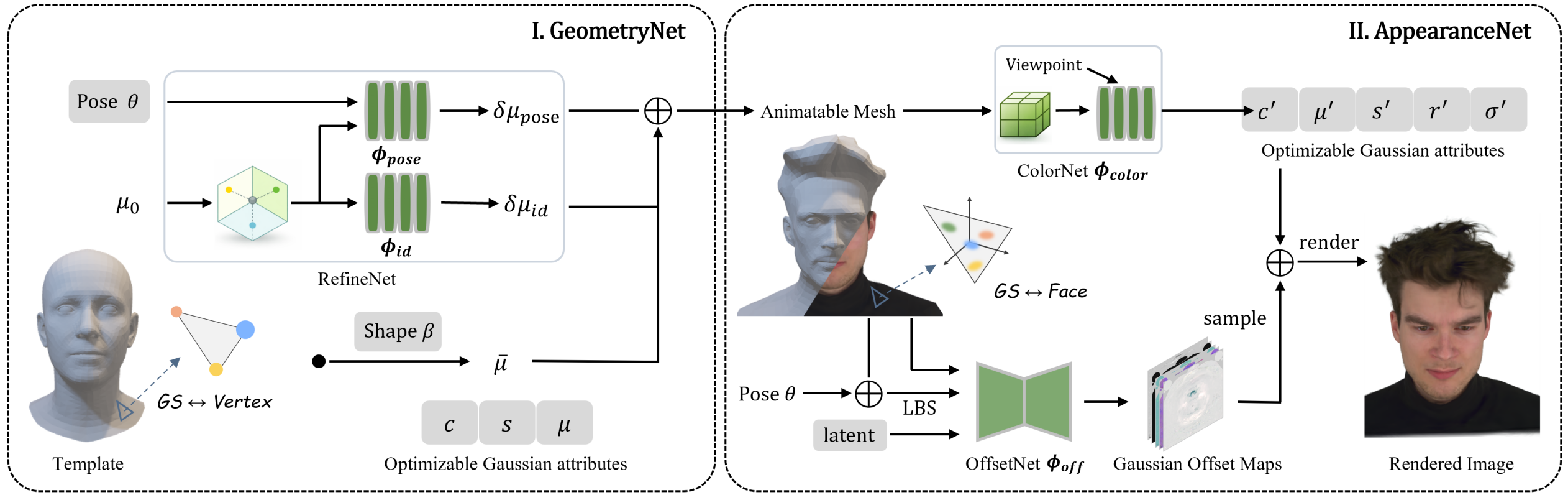}
  \caption{Overview of MGAvatar. We propose an end-to-end framework for high-fidelity head avatar reconstruction with mesh-bound Gaussians. In the geometry stage, FLAME parameters, vertex-bound Gaussians, and a pose-dependent deformation network are jointly optimized to refine identity-specific head geometry. In the appearance stage, face-bound Gaussians and Gaussian attribute offsets model fine-grained appearance, while a grid-based view-conditioned neural field improves color consistency under novel poses and expressions.}
  \label{framework}
\end{figure*}
\section{Method}

Our method models a Gaussian head avatar using the tracked FLAME mesh. The canonical template ${\mathcal{M}}=\{{\mathcal{V}},{\mathcal{F}}\}$ consists of $5K$ vertices and $10K$ faces. The Gaussian properties include  position ${\mu}$, rotation ${r}$, scale ${s}$, opacity ${\sigma}$, and color ${c}$~\cite{kerbl20233d}. As shown in Fig.~\ref{framework}, our method learns the animatable avatar from both geometry and appearance. We first focus on geometry optimization, capturing personalized head features by adjusting the positions of the mesh vertices. The number of faces on the standard template is not enough for intricate geometry adjustment; therefore, we subdivide it to obtain a high-resolution mesh ${\hat{\mathcal{M}}}=\{{\hat{\mathcal{V}}},{\hat{\mathcal{F}}}\}$ with $20K$ vertices and $40K$ faces.

\subsection{Geometry Optimization of Animatable Head Avatar}

The standard FLAME template fails to represent identity-specific facial features and beyond-template structures such as hair and shoulders. To address this limitation, we first investigate how to adjust the mesh geometry. We learn a personalized head structure by optimizing the FLAME parameters and adjusting the mesh vertices. As illustrated in Fig.~\ref{framework}, we optimize FLAME parameters such as 3D pose parameters ${\theta} \in \mathbb{R}^{15}$, shape parameters ${\beta} \in \mathbb{R}^{300}$, and facial expression parameters ${\psi} \in \mathbb{R}^{100}$, to refine personalized geometry and motion. We then introduce a RefineNet to learn mesh vertex offsets to compensate for beyond-template structures and non-rigid deformations.

\textbf{Vertex-bound Gaussians.} We explore a one-to-one binding strategy between Gaussians and the canonical mesh vertices. We initialize the Gaussian centers, denoted as ${\mu_0}$, using the vertex positions of the subdivided mesh ${\hat{\mathcal{M}}}$. To facilitate stable geometric optimization, we reduce the DoF of Gaussian attributes and primarily improve rendering quality through Gaussian position and appearance learning. Under the constraint of regularization terms, we can obtain a gradually and smoothly deformed head mesh.

The rotation ${r}$ of a Gaussian is set to identity matrix to prevent it from reducing rendering error via orientation adjustment. The scale ${s}$ is constrained to be isotropic to avoid elongated shapes that could bypass positional update. The opacity ${\sigma} =1$ is to ensure every Gaussian consistently contributes to gradient propagation, preventing the model from exploiting transparency instead of position for error reduction. The color ${c}$ is represented by the 0th-order component of spherical harmonic (SH), removing view-dependent cues that might otherwise interfere with geometry learning. Therefore, during geometry optimization, the properties of a Gaussian are represented as \( {G_{v}} = \{{\mu}, {s}, {c}, {\sigma}, {r}\} \), where ${\mu}$, ${s}$, and ${c}$ are learnable, while ${\sigma}$ and ${r}$ are fixed. 

\begin{figure}[htb]
    \centering
    \includegraphics[width=1\linewidth]{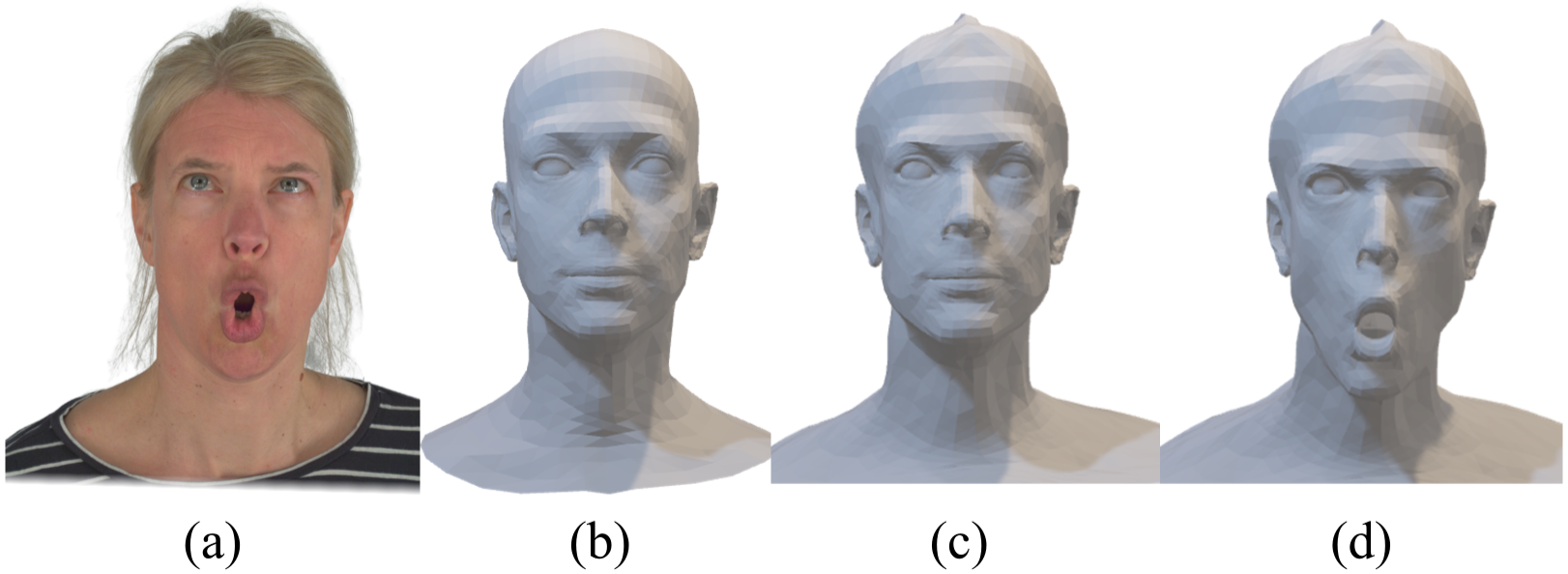}
    \caption{The effectiveness of geometry optimization. (a) one frame; (b) identity-specific FLAME based on refined shapes; (c) animatable head after vertex adjustment; (d) posed head. Geometry optimization is essential for accurate registration of heads.}
    \label{geometry}
\end{figure}

\textbf{Identity-specific geometric refinement.} Our method starts from a neutral FLAME template and refines shape parameters to model the static head geometry of an individual. The canonical Gaussian positions ${\bar{\mu}}$ are then calculated using expression-neutral shape blendshapes, i.e., ${\bar{\mu}} = {\mathcal{T}}\{{\beta}\}$. As shown in Fig.~\ref{geometry}(b), the improved shape perfectly captures face features. Since the geometry of hair and neck/shoulder among individuals cannot be represented by parametric templates, we introduce an identity-specific offset network ${\Phi_{id}}{(\cdot)}$ to model personalized geometry. We take the vertex positions of the mesh as Gaussian positions ${\mu_0}$ and encode them using a triplane ${\mathcal{E}_{tri}}$. Since ${\mu_0}$ remains invariant during training, the triplane features serve as a stable spatial encoding, providing consistent input to the identity-specific offset network ${\Phi_{id}}{(\cdot)}$, thereby enhancing training stability. Then, the canonical Gaussian position ${\mu_c}$ is obtained by combining ${\bar{\mu}}$ with the learned offset:
\begin{equation}
\begin{aligned}
\label{formula: canonical mesh}
    \mu_c &= \bar{\mu} + \delta \mu_{id},
    \delta \mu_{id} = \Phi_{id}(\mathcal{E}_{tri}(\mu_0))
\end{aligned}
\end{equation}

Compared to methods that use UV mapping to predict mesh offsets~\cite{wang2025mega,zhuang2025idol}, our method directly predicts vertex offsets, which can provide greater freedom for geometric adjustments in hair and shoulder regions, while also ensuring the stability of Gaussian binding and training.

\textbf{Pose-dependent geometric refinement.} Head geometry is not static under articulation and exhibits pose-dependent non-rigid deformations induced by neck, jaw, and eye movements. Therefore, we design a pose-dependent geometric transformation network that refines Gaussian positions based on the pose parameters. To ensure that the mesh is closely aligned with the real geometry after linear blend skinning (LBS), both the pose parameters ${\theta}$ and the expression parameters ${\psi}$ are learnable during training and jointly optimized with the deformation network. Pose parameters including neck, jaw, and eye poses are fed together with the triplane-encoded feature ${f}$ into the network ${\Phi_{pose}}{(\cdot)}$ to predict the pose-dependent offsets:
\begin{equation}
\label{formula: pose-dependent offsets}
    \delta \mu_{pose} = \Phi_{pose}(\mathcal{E}_{tri}(\mu_0),\theta_{neck},\theta_{jaw},\theta_{eye}).
\end{equation}
The animatable Gaussian positions are then obtained by combining the updated base positions ${\bar{\mu}}$ with the identity-specific offsets ${\delta \mu_{id}}$ and the pose-dependent offsets ${\delta \mu_{pose}}$:
\begin{equation}
\label{formula: animatable mesh}
    \mu = \bar{\mu} + \delta \mu_{id} + \delta \mu_{pose}.
\end{equation}

As shown in Fig.~\ref{geometry}(c), the animatable head now captures distinctive geometric structures such as hair and shoulders. Then, the animatable mesh and Gaussians are driven by expression and pose parameters to reconstruct the head avatar in observation space.

\textbf{Optimization.} We optimize the model by jointly leveraging color and geometric regularization. Specifically, we measure the MAE and structural similarity between rendering images and ground-truth images.
\begin{equation}
\label{formula: appearance loss1}
    \mathcal{L}_{\text{img}} = \lambda_{\text{c}} \mathcal{L}_{\text{color}} 
                + \lambda_{\text{m}} \mathcal{L}_{\text{ssim}},
\end{equation}
where $\lambda_{\text{c}} = 0.8$, $\lambda_{\text{m}} = 0.2$.

Geometric regularization is crucial in mesh deformation training. Learnable shape parameters have good initial values and do not produce obvious geometric distortions. However, Gaussian-driven vertex position offsets cannot guarantee the smoothness and consistency of mesh changes, which may lead to flipped faces, wrinkles, and discontinuous deformations. Therefore, we constrain the magnitudes of both the identity-specific offsets and pose-dependent offsets predicted by the network. Furthermore, the isotropic Gaussian scale ${s}$ also requires regularization constraints to prevent local instability caused by scales that are too large or too small.
\begin{equation}
\label{formula: offset loss}
    \mathcal{L}_{\text{off}} = \|\delta \mu_{id}\|_2^2 + \|\delta \mu_{pose}\|_2^2 + \|s\|_2^2.
\end{equation}

Moreover, we impose a Laplacian regularization term between the predicted mesh and the FLAME parametric mesh to enforce local consistency of geometric deformation. This encourages the network to learn more refined and stable vertex adjustments with reference to the geometric priors of the template. We also apply the constraint to Gaussian scale to prevent abrupt changes. We define ${\Delta}(\cdot)$ as the Laplacian operator:
\begin{equation}
\label{formula: lap loss}
    \mathcal{L}_{\text{lap}} = \|\Delta(\mu_c)-\Delta(\bar{\mu})\|_2^2 + \|\Delta(\mu)-\Delta(\bar{\mu})\|_2^2
    + \|\Delta(s)\|_2^2.
\end{equation}

The objective function for geometry optimization is defined as:
\begin{equation}
\label{formula: geometric loss}
    \mathcal{L}_{\text{geo}} = \mathcal{L}_{\text{img}}+ \lambda_{\text{o}} \mathcal{L}_{\text{off}} + \lambda_{\text{l}} \mathcal{L}_{\text{lap}},
\end{equation}
where $\lambda_{\text{o}} = 10$, $\lambda_{\text{l}} = 1 \times 10^{5}$.

\subsection{Appearance Optimization Based on Face-Bound Gaussians}

After geometric training stabilizes, our method switches to optimizing rendering quality. We freeze the geometry optimization network and switch the binding strategy from Gaussian-Vertex to Gaussian-Face, allowing Gaussians to float around the triangle faces and decoupling appearance modeling from geometry. We bind the Gaussians to mesh faces with high geometric quality, ensuring that initially each face binds a Gaussian at its center. A local coordinate system is constructed for each triangular face of the mesh. One edge vector, face normal, and their cross product construct a rotation matrix ${R}$ of the triangular face to describe its orientation in global space. The position of the center point of the triangle is ${T}$. A scale factor ${k}$ is computed from edge lengths to capture the overall size of the face. To achieve high-fidelity rendering, we parameterize each Gaussian in a triangle-local coordinate system with position ${\mu^*}$, scale ${s^*}$, and rotation ${r^*}$, where the scale is anisotropic in three dimensions and all other parameters are learnable. Thus the Gaussian in the world coordinate is represented as ${G_{f}} = \{{\mu^\prime}, {s^\prime}, {c^\prime}, {\sigma^\prime}, {r^\prime}\}$. The global position, rotation and scale are converted from local coordinates using the following equations:
\begin{equation}
\label{formula: transform}
\begin{aligned}
    \mu' &= k R \mu^* + T,\\
    r'   &= R r^*,\\
    s'   &= k s^*.
\end{aligned}
\end{equation}
A single Gaussian is often insufficient to cover complex appearance structures, such as curved strands of hair, clothing wrinkles, or fine skin details. Therefore, we employ the adaptive density control strategy to perform Gaussian cloning and splitting, dynamically adjusting their number to accommodate local detail complexity.

\begin{figure*}
    \centering
    \includegraphics[width=1\textwidth]{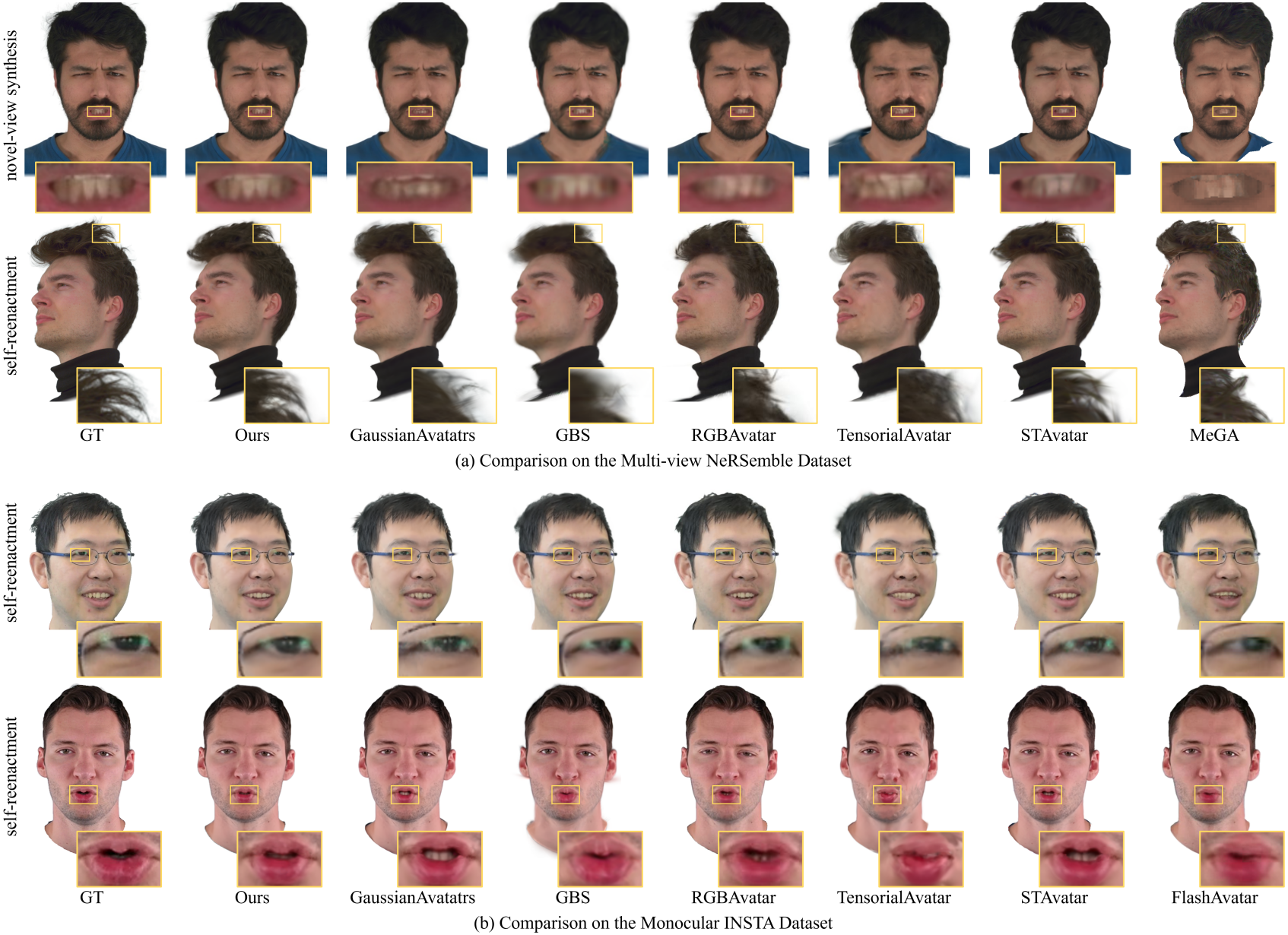}
    \caption{(a) Novel-view synthesis and self-reenactment results on the NeRSemble dataset. (b) Self-reenactment results on the INSTA dataset. Our method better preserves fine details and reduces rendering artifacts, particularly in challenging regions such as hair and teeth.}
    \label{vs}
\end{figure*}

\textbf{Grid-based neural color field.} When the observation views are limited, low-order spherical harmonics struggle to represent complex view-dependent effects, often leading to noticeable artifacts under challenging or unseen poses and expressions. To address this limitation, we introduce ColorNet, a grid-based neural color field to model view-dependent color for Gaussian primitives in a more expressive manner. First, to ensure input stability, we encode the canonical position ${\mu_c}$ of all Gaussians using a Hash Grid ${\mathcal{H}}(\cdot)$ to extract features. Let ${d} \in \mathbb{R}^3$ denote the view direction of the Gaussian relative to the camera. The resulting feature is concatenated with ${d}$ and fed into a lightweight MLP ${\Phi_{color}}(\cdot)$ to produce the view-dependent color ${c'}$ of the Gaussian:
\begin{equation}
\label{formula: color}
    c' = \Phi_{color}\big([\mathcal{H}(\mu_c), d]\big).
\end{equation}

\textbf{Gaussian Offset Net.} The attributes of face-bound Gaussians remain strongly correlated with the underlying mesh motion, limiting their ability to model fine-grained non-rigid facial details, such as skin wrinkles, subtle expression-induced deformations, and high-frequency appearance variations. To address this limitation, we introduce a Gaussian Offset Net that predicts per-Gaussian attribute offsets to enhance the flexibility in Gaussian representations.

Specifically, we construct UV position maps $M_{cano} \in \mathbb{R}^{3 \times H \times W}$ and $M_{pose} \in \mathbb{R}^{3 \times H \times W}$ from the canonical mesh and each posed mesh, respectively. In addition, we introduce a learnable latent $Z \in \mathbb{R}^{3 \times H \times W}$ to capture identity-specific appearance information that cannot be fully described by geometric inputs. 
We design an OffsetNet $\Phi_{off}=\{\Phi_e,\Phi_d\}$ with a convolutional encoder-decoder architecture to predict Gaussian attribute offset maps $O_{map}\in\mathbb{R}^{13\times H\times W}$, including position, color, rotation, scale, and opacity. The three input maps are concatenated and fed into the encoder $\Phi_e(\cdot)$, while the pose parameters $\theta$ are encoded by $\mathcal{E}(\cdot)$ into pose features. These features are then concatenated and fed into the decoder $\Phi_d(\cdot)$ to predict the offset maps. The formulation of OffsetNet is defined as:
\begin{equation}
O_{map}=\Phi_d([\Phi_e([M_{cano},M_{pose},Z]),\mathcal{E}(\theta)]).
\end{equation}

By sampling the offset maps $O_{map}$, we obtain attribute offsets for each face-bound Gaussian. Combining these with the base Gaussian attributes yields the Gaussians for rendering: ${G_{r}} = \{{\mu^\prime+\delta \mu^\prime}, {s^\prime+\delta s^\prime}, {c^\prime+\delta c^\prime}, {\sigma^\prime+\delta \sigma^\prime}, {r^\prime \cdot \delta r^\prime}\}$, where the $\cdot$ operator denotes multiplication of the corresponding rotation matrices.

\begin{table*}[tb]
  \caption{Quantitative comparison on NeRSemble dataset~\cite{kirschstein2023nersemble}. \textbf{Bold} indicates the best and \underline{underline} indicates the second. MeGA and our method optimize the FLAME mesh during training.}
  \label{tab:com1}
  \centering
  \begin{tabular}{@{}l|c|ccc|ccc@{}}
    \toprule
    \multirow{2.5}{*}{Method}
    & \multirow{2.5}{*}{Mesh Optimization}
    & \multicolumn{3}{c}{Novel-View Synthesis}
    & \multicolumn{3}{c}{Self-Reenactment} \\
    \cmidrule(lr){3-5} \cmidrule(lr){6-8}
    & 
    & PSNR $\uparrow$ & SSIM $\uparrow$ & LPIPS $\downarrow$
    & PSNR $\uparrow$ & SSIM $\uparrow$ & LPIPS $\downarrow$ \\
    \midrule
    GaussianAvatars~\cite{qian2024gaussianavatars}         & -- & 31.065 & 0.936 & 0.066 & 25.822 & 0.910 & 0.077 \\
    
    GBS~\cite{ma20243d}                        &    --        & 26.681 & 0.913 & 0.179 & 25.577 & 0.907 & 0.178 \\
    RGBAvatar~\cite{li2025rgbavatar} &         --      &28.919&0.935&0.064&25.815&0.906&0.079\\
    TensorialAvatar~\cite{wang20253d}                       &      --      & 27.339 & 0.914 & 0.097 & 25.885 & 0.909 & 0.100 \\

    MeGA~\cite{wang2025mega}                  & $\checkmark$  & 15.223 & 0.845 & 0.163 & 17.371 & 0.857 & 0.144 \\

    STAvatar~\cite{zhao2026stavatar}                       &     --       & \underline{33.803} & \underline{0.946} & \underline{0.048} & \underline{26.573} & \underline{0.911} & \underline{0.063} \\
    
    Ours                                     & $\checkmark$  & \textbf{34.570} & \textbf{0.953} & \textbf{0.043} & \textbf{26.627} & \textbf{0.913} & \textbf{0.061} \\
    \bottomrule
  \end{tabular}
\end{table*}

\begin{table}[tb]
  \caption{Quantitative comparison on monocular dataset~\cite{gafni2021dynamic,grassal2022neural,zheng2022avatar,zielonka2023instant}. \textbf{Bold} indicates the best and \underline{underline} indicates the second.}
  \label{tab:com2}
  \centering
  \begin{tabular}{@{}l|ccc@{}}
    \toprule
    \multirow{2.5}{*}{Method} 
    & \multicolumn{3}{c}{Self-Reenactment} \\
    \cmidrule(lr){2-4}
    & PSNR $\uparrow$ & SSIM $\uparrow$ & LPIPS $\downarrow$ \\
    \midrule
    FlashAvatar~\cite{xiang2024flashavatar}           &26.817&0.912&0.089\\
    GaussianAvatars~\cite{qian2024gaussianavatars}           &27.544&0.933&0.053\\
    
    GBS~\cite{ma20243d}         &28.346&0.931&0.058\\
    RGBAvatar~\cite{li2025rgbavatar}         &28.399&0.940&0.056\\
    TensorialAvatar~\cite{wang20253d}            &27.179&0.924&0.065\\
    STAvatar~\cite{zhao2026stavatar}
    &\underline{28.874}&\underline{0.941}&\underline{0.040}\\
    Ours&\textbf{29.015}&\textbf{0.943}&\textbf{0.038}\\
    \bottomrule
  \end{tabular}
\end{table}

\textbf{Optimization.} For appearance learning, we apply color and SSIM supervision to the rendered image. During optimization, Gaussians may deviate from their bounded faces, leading to rendering artifacts under novel poses or expressions. Gaussian with relatively large size related to its parent triangle are very sensitive to mesh motion, and easily introduce jitter and artifacts when rotating along with the mesh. Thus we regularize the position and scale of the Gaussian and grant it a degree of freedom using thresholds. In addition, we design a regularization term to constrain Gaussian attribute offsets. The training objective function is as follows:

\begin{equation}
\label{formula: appearance loss2}
\begin{aligned}
\mathcal{L}_{\text{app}}
=&\ \lambda_{\text{c}} \mathcal{L}_{\text{color}} + \lambda_{\text{m}} \mathcal{L}_{\text{ssim}} + \lambda_{\mu}\| \max ( \mu- \epsilon_{\mu}, 0 ) \|_2\\
+& \lambda_{s}\| \max ( s- \epsilon_{s}, 0 ) \|_2 + \lambda_{\text{a}} \| O_{map} \|_2.
\end{aligned}
\end{equation}

Here, $\epsilon_{\mu}$ is set to $1$ to constrain the Gaussian to move around the corresponding face, $\epsilon_{s}$ is set to $0.6$ to allow the Gaussian to change shape with a certain degree of freedom, and the corresponding weights are set to $\lambda_{\text{c}}=0.8$, $\lambda_{\text{m}}=0.2$, $\lambda_{\mu}=0.01$, $\lambda_{s}=1$,  $\lambda_{\text{a}}=0.01$.

\section{Experiments}

\subsection{Datasets and Baselines}

We evaluate our method on both multi-view and monocular video datasets. The multi-view NeRSemble dataset~\cite{kirschstein2023nersemble} contains 11 sequences per subject captured from 16 viewpoints, including 10 guided expression sequences and one free-form performance. For monocular evaluation, we use 10 videos from INSTA~\cite{zielonka2023instant}, NHA~\cite{grassal2022neural}, IMAvatar~\cite{zheng2022avatar}, and NerFace~\cite{gafni2021dynamic}. On NeRSemble, we evaluate novel-view synthesis and self-reenactment using same-identity driving expressions. For the monocular datasets, we evaluate self-reenactment results. We additionally present cross-identity reenactment results driven by different identities to validate the robustness of our method.

We compare with state-of-the-art 3DGS-based head avatar methods. For both multi-view and monocular settings, we compare with GaussianAvatars~\cite{qian2024gaussianavatars}, GaussianBlendShape (GBS)~\cite{ma20243d}, RGBAvatar~\cite{li2025rgbavatar}, TensorialAvatar~\cite{wang20253d}, and STAvatar~\cite{zhao2026stavatar}. Additionally, MeGA~\cite{wang2025mega} is included only in the multi-view setting due to its focus on multi-view reconstruction, while FlashAvatar~\cite{xiang2024flashavatar} is included only in the monocular setting as it is specifically designed for monocular head avatar reconstruction.

\subsection{Implementation Details}
To ensure a fair comparison, all experiments are conducted on a single NVIDIA RTX 4090 GPU. The total number of training iterations in our method is set to 340k. Among them, the first 100k iterations are dedicated to geometry optimization for learning a personalized head mesh. The learning rate of RefineNet is set to $\eta_{\text{geo}} = 1 \times 10^{-3}$. Once the geometric structure stabilizes, we freeze the RefineNet parameters and shift the training focus toward appearance learning, where the learning rates of the ColorNet and OffsetNet are set to $\eta_{\text{color}} = 1 \times 10^{-2}$ and $\eta_{\text{offset}} = 1 \times 10^{-4}$, respectively.

The identity-specific offset network $\Phi_{id}$ and pose-dependent offset network $\Phi_{pose}$ are 4-layer 128-channel MLPs. ColorNet $\Phi_{color}$ employs hash grids containing 2-channel features at 16 different resolutions (ranging from 16 to 4096), supplemented by a subsequent 2-layer 64-channel MLP. OffsetNet $\Phi_{off}$ consists of a lightweight encoder $\Phi_e$ and a decoder $\Phi_d$. The encoder $\Phi_e$ comprises an 8-layer convolutional feature extractor and a single-layer UV position map encoder, while the decoder $\Phi_d$ consists of 5 convolutional layers.

\begin{figure*}
    \centering    
    \includegraphics[width=1\textwidth]{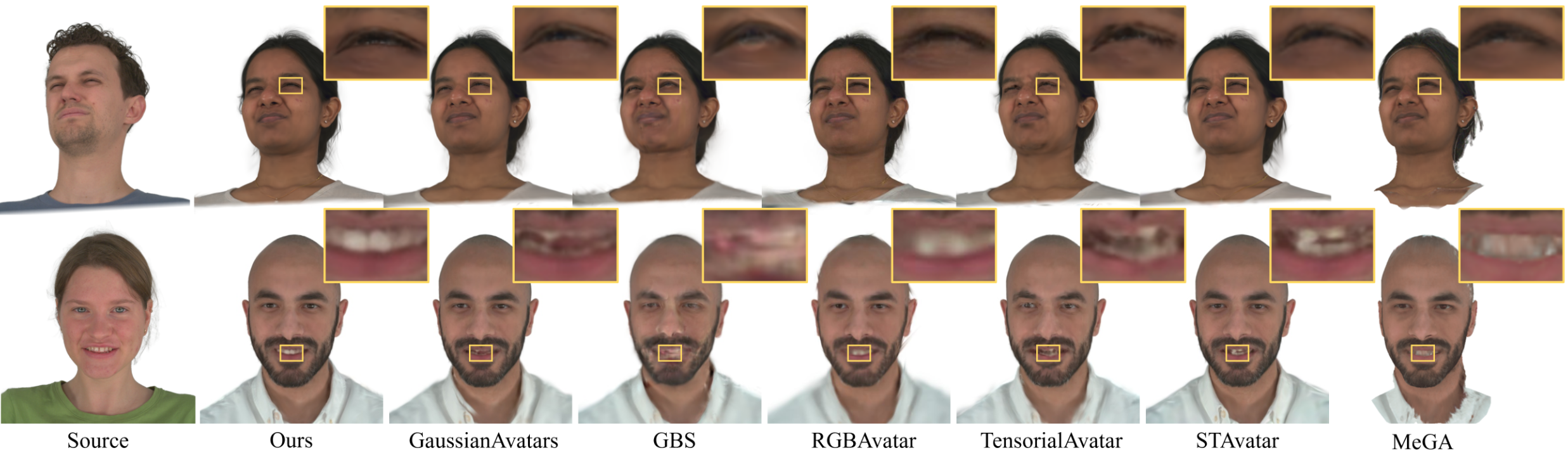}
    \caption{Cross-identity reenactment results. Our method transfers expressions while preserving identity-specific geometry and appearance.}
    \label{crossid}
\end{figure*}

\begin{figure}
    \centering
    \includegraphics[width=1\linewidth]{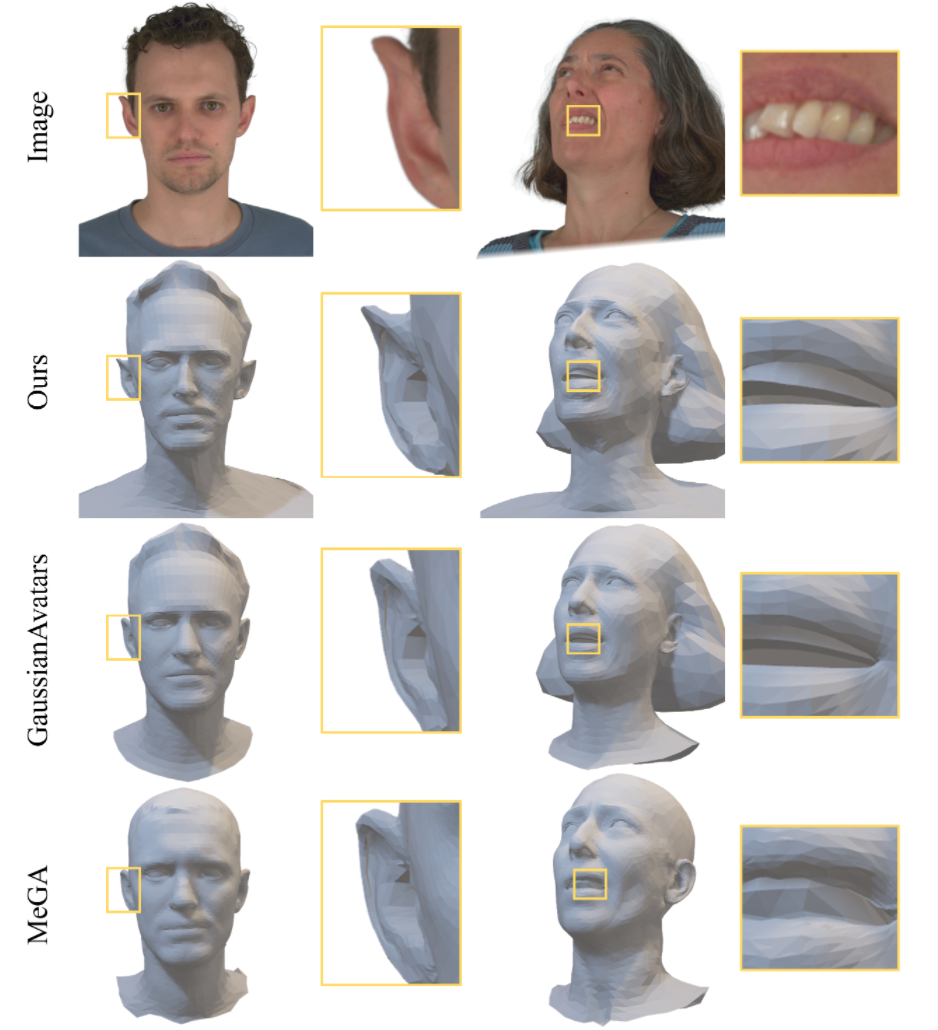}
    \caption{Optimized mesh results on the NeRSemble dataset. GaussianAvatars uses fixed VHAP-pretrained vertex offsets, while MeGA lacks explicit hair modeling.}
    \label{meshvs1}
\end{figure}
\subsection{Comparative Experiments}
\textbf{Quantitative Results.} We report PSNR, SSIM, and LPIPS evaluation metrics on NeRSemble and monocular datasets in Tab.~\ref{tab:com1} and Tab.~\ref{tab:com2}, respectively. On the NeRSemble dataset, MeGA and our method optimize mesh vertex positions during training, whereas other methods—such as RGBAvatar, TensorialAvatar, and STAvatar—lack the capability to learn personalized meshes and instead utilize meshes preprocessed using the VHAP~\cite{qian2024vhap} method proposed by GaussianAvatars as input. Our method outperforms other methods in both novel-view synthesis and self-reenactment tasks. All evaluation metrics are computed over the full image. MeGA achieves relatively poor quantitative results due to its inability to model the shoulder region properly. 
The improved rendering quality mainly stems from more accurate underlying geometry for appearance modeling, a neural color field for better cross-view consistency, and OffsetNet for more flexible Gaussian representations of high-frequency details.

\textbf{Qualitative Results.} In terms of rendering quality, our method achieves superior visual fidelity. 
As illustrated in Fig.~\ref{vs}, it produces more accurate facial and clothing details on the multi-view dataset, benefiting from refined facial and shoulder meshes that provide reliable priors for face-bound Gaussians. In challenging unseen views, existing methods often fail to reconstruct accurate tooth structures, whereas our method generates clearer and more geometrically consistent teeth.
Self-reenactment results on both multi-view and monocular datasets further demonstrate that MGAvatar maintains strong geometric and appearance consistency under novel poses, producing fine hair strands and more accurate pose transfer.
GaussianAvatars, GBS, and TensorialAvatar produce blurry hair and lose fine details. RGBAvatar and STAvatar introduce noticeable hair artifacts caused by elongated Gaussians. MeGA exhibits visible gaps between the hair and face regions, leading to obvious stitching artifacts. Some methods also suffer from inaccurate pose transfer, particularly TensorialAvatar.
To further demonstrate the quality of our method, we conduct several cross-identity reenactment experiments. As shown in  Fig.\ref{crossid}, we use the expression and pose of a source identity to drive our reconstructed heads. The results indicate that our method can faithfully transfer expressions while preserving identity-specific geometry and appearance.

\textbf{Optimized Mesh.} To further demonstrate the geometric advantages of our method, we compare the reconstructed meshes with those produced by GaussianAvatars and MeGA in Fig.~\ref{meshvs1}. GaussianAvatars relies on a pre-optimized personalized mesh by VHAP~\cite{qian2024vhap} and does not further adjust vertices during Gaussian optimization, while MeGA predicts facial vertex displacements and represents hair using separate Gaussian primitives without an explicit mesh representation. In contrast, our method directly optimizes mesh vertices under image-based rendering supervision, enabling personalized geometric refinement during training. Moreover, our geometry learning stage converges in approximately 30 minutes, whereas GaussianAvatars requires more than 3 hours of mesh preprocessing per subject and MeGA involves an even more time-consuming geometry optimization process. 

As shown in Fig.~\ref{meshvs1}, our method reconstructs a complete personalized head mesh, including the face, hair, ears, and shoulders. This is achieved through vertex-level optimization, which allows the mesh to deform beyond the neutral template and generate shoulder regions absent in the initial mesh. Such capability is not supported by either the preprocessing of GaussianAvatars or the optimization in MeGA. Furthermore, as illustrated in Fig.~\ref{geometry}(d), the reconstructed mesh preserves the original FLAME topology, enabling direct compatibility with FLAME-based control and animation.

\begin{figure}
    \centering
    \includegraphics[width=1\linewidth]{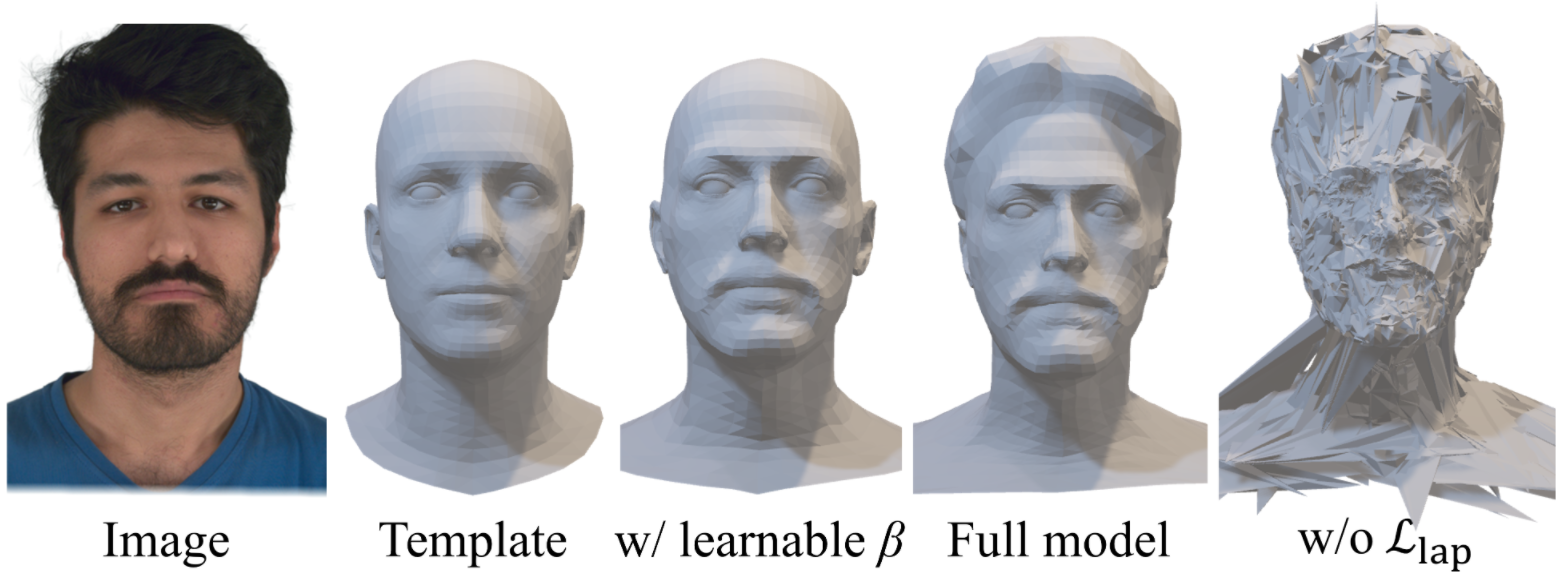}
    \caption{Ablation on geometry optimization. Learnable shape parameters $\beta$ deform the template to capture identity-specific facial geometry, while vertex offsets $\delta \mu_{id}$ and $\delta \mu_{pose}$ further refine the mesh to model hair and shoulder regions; the Laplacian regularization $\mathcal{L}_{\text{lap}}$ enforces mesh smoothness and structural stability.}
    \label{geo}
\end{figure}

\begin{table}[tb]
  \caption{Quantitative ablation on NeRSemble dataset~\cite{kirschstein2023nersemble}. \textbf{Bold} indicates the best.}
  \label{tab:ab}
  \centering
  \setlength{\tabcolsep}{0.6pt}
  \begin{tabular}{@{}l|ccc|ccc@{}}
    \toprule
    \multirow{2.5}{*}{Method} 
    & \multicolumn{3}{c}{Novel-View Synthesis} 
    & \multicolumn{3}{c}{Self-Reenactment} \\
    \cmidrule(lr){2-4} \cmidrule(lr){5-7}
    & PSNR $\uparrow$ & SSIM $\uparrow$ & LPIPS $\downarrow$
    & PSNR $\uparrow$ & SSIM $\uparrow$ & LPIPS $\downarrow$ \\
    \midrule
    w/o $\delta \mu_{pose}$                &33.436&0.945&0.056&26.059&0.906&0.068\\
    w/o $\mathcal{L}_{\text{lap}}$        &34.121&0.948&0.057&25.833&0.898&0.073\\
    w/o ColorNet  &34.193&0.947&0.053&26.129&0.902&0.070\\
    w/o OffsetNet               &32.064&0.941&0.061&26.206&0.911&0.072\\
    Ours                                    & \textbf{34.570} & \textbf{0.953} & \textbf{0.043} & \textbf{26.627} & \textbf{0.913} & \textbf{0.061} \\
    \bottomrule
  \end{tabular}
\end{table}

\subsection{Ablation Study}
\begin{figure}
    \centering
    \includegraphics[width=1\linewidth]{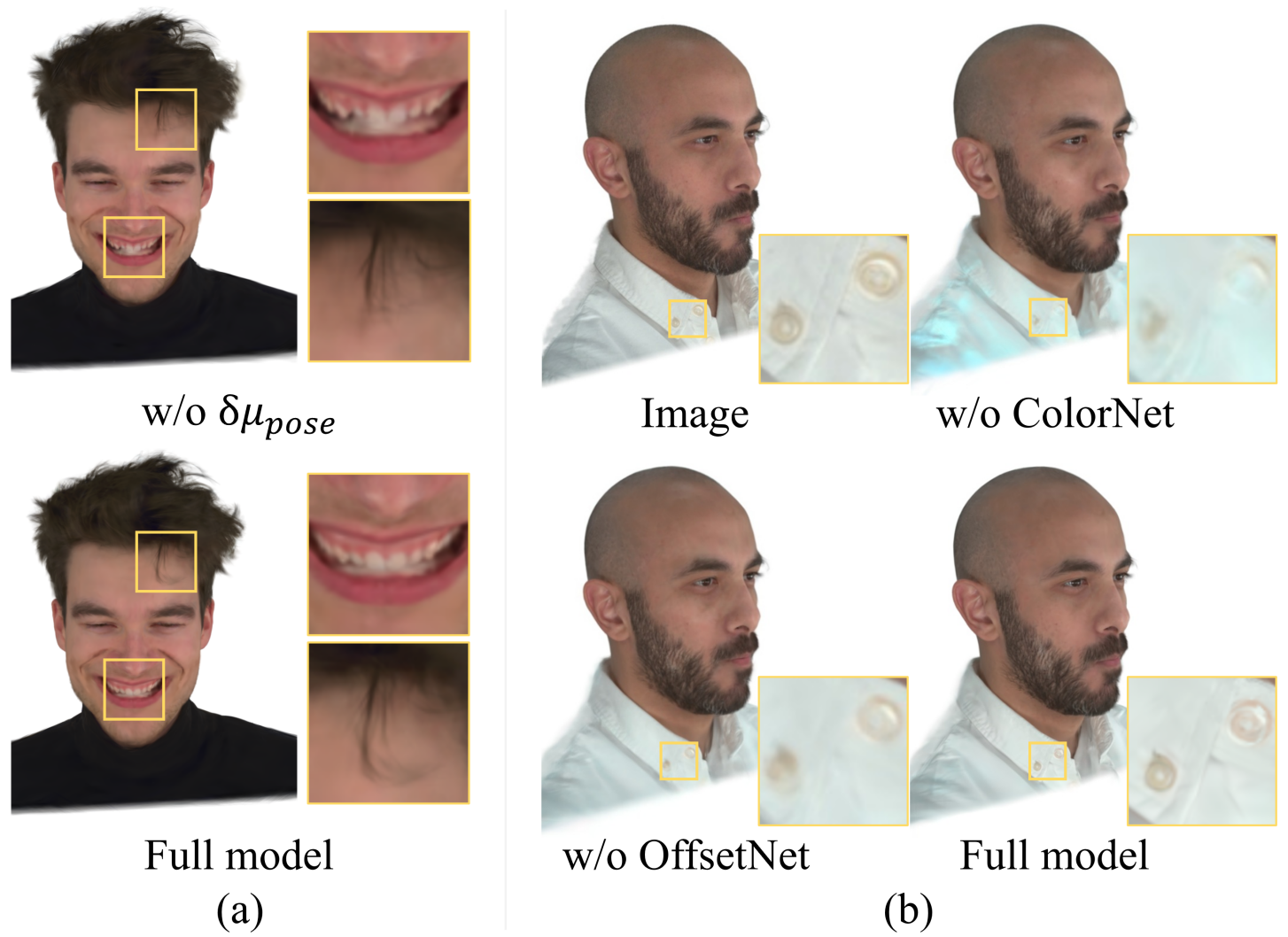}
    \caption{(a) Ablation on geometry optimization. (b) Ablation on appearance optimization. }
    \label{ab}
\end{figure}

To systematically evaluate the contribution of each component in MGAvatar, we conduct a comprehensive ablation study. Quantitative results are shown in Tab.~\ref{tab:ab}, while qualitative comparisons are illustrated in Fig.~\ref{geo} and Fig.~\ref{ab}. Tab.~\ref{tab:ab} shows that removing each module reduces rendering performance on tasks with novel views and novel expressions.


\textbf{Ablation on geometry optimization.} In geometry learning, we optimize the FLAME shape parameters and predict pose-dependent vertex offsets $\delta \mu_{pose}$ to obtain a personalized animatable mesh. As shown in Fig.~\ref{geo}, starting from a neutral FLAME template, the learned shape parameters recover identity-specific facial features. In addition, $\delta \mu_{pose}$ captures geometric deformations caused by pose. As shown in the first row of Tab.~\ref{tab:ab}, removing $\delta \mu_{pose}$ leads to a drop in performance. Moreover, $\delta \mu_{pose}$ improves rendering quality. As shown in Fig.~\ref{ab}(a), removing it leads to the loss of facial details. Laplacian regularization is crucial for geometric stability. Removing $\mathcal{L}_{\text{lap}}$ significantly degrades mesh quality. As shown in the last column of Fig.~\ref{geo}, the resulting mesh exhibits severe geometric artifacts, including flipped triangles, irregular face scales, and broken surface connectivity.



\textbf{Ablation on appearance optimization.} 
As shown in Fig.~\ref{ab}(b), SH-based representation introduces noticeable blue artifacts on white clothing regions. In contrast, the neural color field better captures view-dependent color variations, improving appearance fidelity.
For the OffsetNet, results in Tab.~\ref{tab:ab} show a clear performance drop when it is removed. The OffsetNet provides dynamic adjustment for face-bound Gaussians, enabling better adaptation to expressions and poses and recovering high-frequency details, such as clothing wrinkles, as shown in Fig.~\ref{ab}(b).


\subsection{Limitations and Future Works} 
MGAvatar achieves significant improvements in geometry and appearance modeling. Built on the standard FLAME template, the method does not modify the topology but optimizes the shape through vertex-level adjustments. As a result, its ability to model complex structures, such as hair strands and clothing wrinkles, is limited.
In addition, since the FLAME template lacks complete oral structures and monocular supervision provides weak geometric constraints, the mouth reconstruction accuracy still has room for improvement. In future work, we plan to explore more flexible 3D representations to better model hair, clothing, and oral structures.


\section{Conclusion}
We present MGAvatar, a mesh-bound Gaussian representation for high-fidelity animatable head avatars. By introducing two mesh–Gaussian binding strategies with different degrees of freedom, MGAvatar effectively decouples geometry and appearance optimization. Through joint optimization of parametric templates, mesh deformation, ColorNet, and OffsetNet, our method learns personalized geometry that is fully compatible with FLAME-based driving, while producing high-fidelity appearance. Extensive experiments demonstrate that MGAvatar achieves superior visual quality and generalization ability in both novel-view synthesis and reenactment tasks.
\section*{Acknowledgments}
This work is supported in part by Guangdong Higher Education Upgrading Plan (No. 2024KTSCX223) at Beijing Normal-Hong Kong Baptist University, in part by BNBU Centre for Computational Culture and Heritage \textbar\ NVIDIA DLI, and in part by Guangdong and Hong Kong Universities ``1+1+1” Joint Research Collaboration Scheme (No. 2025A0505000003).

\bibliographystyle{eg-alpha-doi} 
\bibliography{egbibsample}       

\newcommand{\etalchar}[1]{$^{#1}$}
\begin{thebibliography}{\uppercase{GMFB{\etalchar{*}}18}}

\bibitem[ASS23]{athar2023flame}
\textsc{Athar S., Shu Z., Samaras D.}:
\newblock Flame-in-nerf: Neural control of radiance fields for free view face animation.
\newblock In \emph{2023 IEEE 17th International Conference on Automatic Face and Gesture Recognition (FG)} (2023), IEEE, pp.~1--8.

\bibitem[BCLT21]{bai2021riggable}
\textsc{Bai Z., Cui Z., Liu X., Tan P.}:
\newblock Riggable 3d face reconstruction via in-network optimization.
\newblock In \emph{Proceedings of the IEEE/CVF Conference on Computer Vision and Pattern Recognition} (2021), pp.~6216--6225.

\bibitem[BTH{\etalchar{*}}23]{bai2023learning}
\textsc{Bai Z., Tan F., Huang Z., Sarkar K., Tang D., Qiu D., Meka A., Du R., Dou M., Orts-Escolano S., et~al.}:
\newblock Learning personalized high quality volumetric head avatars from monocular rgb videos.
\newblock In \emph{Proceedings of the IEEE/CVF Conference on Computer Vision and Pattern Recognition} (2023), pp.~16890--16900.

\bibitem[BV23]{blanz2023morphable}
\textsc{Blanz V., Vetter T.}:
\newblock A morphable model for the synthesis of 3d faces.
\newblock In \emph{Seminal Graphics Papers: Pushing the Boundaries, Volume 2}. ACM New York, NY, USA, 2023, pp.~157--164.

\bibitem[BZH{\etalchar{*}}23]{bharadwaj2023flare}
\textsc{Bharadwaj S., Zheng Y., Hilliges O., Black M.~J., Fernandez-Abrevaya V.}:
\newblock Flare: Fast learning of animatable and relightable mesh avatars.
\newblock \emph{arXiv Preprint arXiv:2310.17519} (2023).

\bibitem[CVSW20]{chaudhuri2020personalized}
\textsc{Chaudhuri B., Vesdapunt N., Shapiro L., Wang B.}:
\newblock Personalized face modeling for improved face reconstruction and motion retargeting.
\newblock In \emph{European Conference on Computer Vision} (2020), Springer, pp.~142--160.

\bibitem[CWL{\etalchar{*}}24]{chen2024monogaussianavatar}
\textsc{Chen Y., Wang L., Li Q., Xiao H., Zhang S., Yao H., Liu Y.}:
\newblock Monogaussianavatar: Monocular gaussian point-based head avatar.
\newblock In \emph{ACM SIGGRAPH 2024 Conference Papers} (2024), pp.~1--9.

\bibitem[CWZ{\etalchar{*}}13]{cao2013facewarehouse}
\textsc{Cao C., Weng Y., Zhou S., Tong Y., Zhou K.}:
\newblock Facewarehouse: A 3d facial expression database for visual computing.
\newblock \emph{IEEE Transactions on Visualization and Computer Graphics 20}, 3 (2013), 413--425.

\bibitem[CXW{\etalchar{*}}24]{cai2024hybrid}
\textsc{Cai H., Xiao Y., Wang X., Li J., Guo Y., Fan Y., Gao S., Zhang J.}:
\newblock Hybrid explicit representation for ultra-realistic head avatars.
\newblock \emph{arXiv preprint arXiv:2403.11453} (2024).

\bibitem[DD07]{demirel2007applications}
\textsc{Demirel H.~O., Duffy V.~G.}:
\newblock Applications of digital human modeling in industry.
\newblock In \emph{International Conference on Digital Human Modeling} (2007), Springer, pp.~824--832.

\bibitem[DNM{\etalchar{*}}24]{dhamo2024headgas}
\textsc{Dhamo H., Nie Y., Moreau A., Song J., Shaw R., Zhou Y., P{\'e}rez-Pellitero E.}:
\newblock Headgas: Real-time animatable head avatars via 3d gaussian splatting.
\newblock In \emph{European Conference on Computer Vision} (2024), Springer, pp.~459--476.

\bibitem[DWS{\etalchar{*}}23]{duan2023bakedavatar}
\textsc{Duan H.-B., Wang M., Shi J.-C., Chen X.-C., Cao Y.-P.}:
\newblock Bakedavatar: Baking neural fields for real-time head avatar synthesis.
\newblock \emph{ACM Transactions on Graphics (ToG) 42}, 6 (2023), 1--17.

\bibitem[FLB{\etalchar{*}}23]{feng2023learning}
\textsc{Feng Y., Liu W., Bolkart T., Yang J., Pollefeys M., Black M.~J.}:
\newblock Learning disentangled avatars with hybrid 3d representations.
\newblock \emph{arXiv Preprint arXiv:2309.06441} (2023).

\bibitem[GMFB{\etalchar{*}}18]{gerig2018morphable}
\textsc{Gerig T., Morel-Forster A., Blumer C., Egger B., Luthi M., Sch{\"o}nborn S., Vetter T.}:
\newblock Morphable face models-an open framework.
\newblock In \emph{2018 13th IEEE International Conference on Automatic Face \& Gesture Recognition (FG 2018)} (2018), IEEE, pp.~75--82.

\bibitem[GPL{\etalchar{*}}22]{grassal2022neural}
\textsc{Grassal P.-W., Prinzler M., Leistner T., Rother C., Nie{\ss}ner M., Thies J.}:
\newblock Neural head avatars from monocular rgb videos.
\newblock In \emph{Proceedings of the IEEE/CVF Conference on Computer Vision and Pattern Recognition} (2022), pp.~18653--18664.

\bibitem[GTZN21]{gafni2021dynamic}
\textsc{Gafni G., Thies J., Zollhofer M., Nie{\ss}ner M.}:
\newblock Dynamic neural radiance fields for monocular 4d facial avatar reconstruction.
\newblock In \emph{Proceedings of the IEEE/CVF Conference on Computer Vision and Pattern Recognition} (2021), pp.~8649--8658.

\bibitem[GZX{\etalchar{*}}22]{gao2022reconstructing}
\textsc{Gao X., Zhong C., Xiang J., Hong Y., Guo Y., Zhang J.}:
\newblock Reconstructing personalized semantic facial nerf models from monocular video.
\newblock \emph{ACM Transactions on Graphics (TOG) 41}, 6 (2022), 1--12.

\bibitem[HDP20]{he2020collabovr}
\textsc{He Z., Du R., Perlin K.}:
\newblock Collabovr: A reconfigurable framework for creative collaboration in virtual reality.
\newblock In \emph{2020 IEEE International Symposium on Mixed and Augmented Reality (ISMAR)} (2020), IEEE, pp.~542--554.

\bibitem[HPX{\etalchar{*}}22]{hong2022headnerf}
\textsc{Hong Y., Peng B., Xiao H., Liu L., Zhang J.}:
\newblock Headnerf: A real-time nerf-based parametric head model.
\newblock In \emph{Proceedings of the IEEE/CVF Conference on Computer Vision and Pattern Recognition} (2022), pp.~20374--20384.

\bibitem[HZZ{\etalchar{*}}24]{hu2024gaussianavatar}
\textsc{Hu L., Zhang H., Zhang Y., Zhou B., Liu B., Zhang S., Nie L.}:
\newblock Gaussianavatar: Towards realistic human avatar modeling from a single video via animatable 3d gaussians.
\newblock In \emph{Proceedings of the IEEE/CVF Conference on Computer Vision and Pattern Recognition} (2024), pp.~634--644.

\bibitem[JHX{\etalchar{*}}25]{jin2025sfdm}
\textsc{Jin D., Hu J., Xu B., Dai Y., Qian C., He Y.}:
\newblock Sfdm: Robust decomposition of geometry and reflectance for realistic face rendering from sparse-view images.
\newblock In \emph{2025 IEEE/CVF Conference on Computer Vision and Pattern Recognition (CVPR)} (2025), IEEE, pp.~26409--26419.

\bibitem[KKL{\etalchar{*}}23]{kerbl20233d}
\textsc{Kerbl B., Kopanas G., Leimk{\"u}hler T., Drettakis G., et~al.}:
\newblock 3d gaussian splatting for real-time radiance field rendering.
\newblock \emph{ACM Trans. Graph. 42}, 4 (2023), 139--1.

\bibitem[KQG{\etalchar{*}}23]{kirschstein2023nersemble}
\textsc{Kirschstein T., Qian S., Giebenhain S., Walter T., Nie{\ss}ner M.}:
\newblock Nersemble: Multi-view radiance field reconstruction of human heads.
\newblock \emph{ACM Transactions on Graphics (TOG) 42}, 4 (2023), 1--14.

\bibitem[KSLZ22]{khakhulin2022realistic}
\textsc{Khakhulin T., Sklyarova V., Lempitsky V., Zakharov E.}:
\newblock Realistic one-shot mesh-based head avatars.
\newblock In \emph{European Conference on Computer Vision} (2022), Springer, pp.~345--362.

\bibitem[LBB{\etalchar{*}}17]{li2017learning}
\textsc{Li T., Bolkart T., Black M.~J., Li H., Romero J.}:
\newblock Learning a model of facial shape and expression from 4d scans.
\newblock \emph{ACM Trans. Graph. 36}, 6 (2017), 194--1.

\bibitem[LLW{\etalchar{*}}25]{li2025rgbavatar}
\textsc{Li L., Li Y., Weng Y., Zheng Y., Zhou K.}:
\newblock Rgbavatar: Reduced gaussian blendshapes for online modeling of head avatars.
\newblock In \emph{Proceedings of the Computer Vision and Pattern Recognition Conference} (2025), pp.~10747--10757.

\bibitem[LMR{\etalchar{*}}23]{loper2023smpl}
\textsc{Loper M., Mahmood N., Romero J., Pons-Moll G., Black M.~J.}:
\newblock Smpl: A skinned multi-person linear model.
\newblock In \emph{Seminal Graphics Papers: Pushing the Boundaries, Volume 2}. ACM New York, NY, USA, 2023, pp.~851--866.

\bibitem[LSS{\etalchar{*}}21]{lombardi2021mixture}
\textsc{Lombardi S., Simon T., Schwartz G., Zollhoefer M., Sheikh Y., Saragih J.}:
\newblock Mixture of volumetric primitives for efficient neural rendering.
\newblock \emph{ACM Transactions on Graphics (ToG) 40}, 4 (2021), 1--13.

\bibitem[LSSS18]{lombardi2018deep}
\textsc{Lombardi S., Saragih J., Simon T., Sheikh Y.}:
\newblock Deep appearance models for face rendering.
\newblock \emph{ACM Transactions on Graphics (ToG) 37}, 4 (2018), 1--13.

\bibitem[LXL{\etalchar{*}}25]{liao2025hhavatar}
\textsc{Liao Z., Xu Y., Li Z., Li Q., Zhou B., Bai R., Xu D., Zhang H., Liu Y.}:
\newblock Hhavatar: Gaussian head avatar with dynamic hairs.
\newblock \emph{IEEE Transactions on Pattern Analysis and Machine Intelligence} (2025).

\bibitem[MSS{\etalchar{*}}21]{ma2021pixel}
\textsc{Ma S., Simon T., Saragih J., Wang D., Li Y., De~La~Torre F., Sheikh Y.}:
\newblock Pixel codec avatars.
\newblock In \emph{Proceedings of the IEEE/CVF Conference on Computer Vision and Pattern Recognition} (2021), pp.~64--73.

\bibitem[MSS24]{moon2024expressive}
\textsc{Moon G., Shiratori T., Saito S.}:
\newblock Expressive whole-body 3d gaussian avatar.
\newblock In \emph{European Conference on Computer Vision} (2024), Springer, pp.~19--35.

\bibitem[MST{\etalchar{*}}21]{mildenhall2021nerf}
\textsc{Mildenhall B., Srinivasan P.~P., Tancik M., Barron J.~T., Ramamoorthi R., Ng R.}:
\newblock Nerf: Representing scenes as neural radiance fields for view synthesis.
\newblock \emph{Communications of the ACM 65}, 1 (2021), 99--106.

\bibitem[MWSZ24]{ma20243d}
\textsc{Ma S., Weng Y., Shao T., Zhou K.}:
\newblock 3d gaussian blendshapes for head avatar animation.
\newblock In \emph{ACM SIGGRAPH 2024 Conference Papers} (2024), pp.~1--10.

\bibitem[PCG{\etalchar{*}}19]{pavlakos2019expressive}
\textsc{Pavlakos G., Choutas V., Ghorbani N., Bolkart T., Osman A.~A., Tzionas D., Black M.~J.}:
\newblock Expressive body capture: 3d hands, face, and body from a single image.
\newblock In \emph{Proceedings of the IEEE/CVF Conference on Computer Vision and Pattern Recognition} (2019), pp.~10975--10985.

\bibitem[PFM{\etalchar{*}}25]{peng2025genericavatar}
\textsc{Peng S., Fu Y., Miu R., Lv T., Yang B., Dong X.}:
\newblock Genericavatar: generic human modeling from monocular video based on mesh-guided gaussians: S. peng et al.
\newblock \emph{The Visual Computer 41}, 9 (2025), 6657--6670.

\bibitem[PVO{\etalchar{*}}20]{ploumpis2020towards}
\textsc{Ploumpis S., Ververas E., O'Sullivan E., Moschoglou S., Wang H., Pears N., Smith W.~A., Gecer B., Zafeiriou S.}:
\newblock Towards a complete 3d morphable model of the human head.
\newblock \emph{IEEE Transactions on Pattern Analysis and Machine Intelligence 43}, 11 (2020), 4142--4160.

\bibitem[PXW{\etalchar{*}}25]{peng2025rmavatar}
\textsc{Peng S., Xie W., Wang Z., Guo X., Chen Z., Yang B., Dong X.}:
\newblock Rmavatar: Photorealistic human avatar reconstruction from monocular video based on rectified mesh-embedded gaussians.
\newblock \emph{Graphical Models 139} (2025), 101266.

\bibitem[Qia24]{qian2024vhap}
\textsc{Qian S.}:
\newblock Vhap: Versatile head alignment with adaptive appearance priors.
\newblock \emph{Zenodo} (2024).

\bibitem[QKS{\etalchar{*}}24]{qian2024gaussianavatars}
\textsc{Qian S., Kirschstein T., Schoneveld L., Davoli D., Giebenhain S., Nie{\ss}ner M.}:
\newblock Gaussianavatars: Photorealistic head avatars with rigged 3d gaussians.
\newblock In \emph{Proceedings of the IEEE/CVF Conference on Computer Vision and Pattern Recognition} (2024), pp.~20299--20309.

\bibitem[RFD{\etalchar{*}}24]{retsinas20243d}
\textsc{Retsinas G., Filntisis P.~P., Danecek R., Abrevaya V.~F., Roussos A., Bolkart T., Maragos P.}:
\newblock 3d facial expressions through analysis-by-neural-synthesis.
\newblock In \emph{Proceedings of the IEEE/CVF Conference on Computer Vision and Pattern Recognition} (2024), pp.~2490--2501.

\bibitem[SWL{\etalchar{*}}24]{shao2024splattingavatar}
\textsc{Shao Z., Wang Z., Li Z., Wang D., Lin X., Zhang Y., Fan M., Wang Z.}:
\newblock Splattingavatar: Realistic real-time human avatars with mesh-embedded gaussian splatting.
\newblock In \emph{Proceedings of the IEEE/CVF Conference on Computer Vision and Pattern Recognition} (2024), pp.~1606--1616.

\bibitem[SWX{\etalchar{*}}25]{sun2025svg}
\textsc{Sun H., Wang C., Xu T.-X., Huang J., Kang D., Guo C., Zhang S.-H.}:
\newblock Svg-head: Hybrid surface-volumetric gaussians for high-fidelity head reconstruction and real-time editing.
\newblock In \emph{2025 IEEE/CVF International Conference on Computer Vision (ICCV)} (2025), IEEE, pp.~13326--13335.

\bibitem[TL18]{tran2018nonlinear}
\textsc{Tran L., Liu X.}:
\newblock Nonlinear 3d face morphable model.
\newblock In \emph{Proceedings of the IEEE Conference on Computer Vision and Pattern Recognition} (2018), pp.~7346--7355.

\bibitem[Wag09]{waggoner2009my}
\textsc{Waggoner Z.}:
\newblock \emph{My avatar, my self: Identity in video role-playing games}.
\newblock McFarland, 2009.

\bibitem[WKC{\etalchar{*}}23]{wang2023neural}
\textsc{Wang C., Kang D., Cao Y.-P., Bao L., Shan Y., Zhang S.-H.}:
\newblock Neural point-based volumetric avatar: Surface-guided neural points for efficient and photorealistic volumetric head avatar.
\newblock In \emph{SIGGRAPH Asia 2023 Conference Papers} (2023), pp.~1--12.

\bibitem[WKS{\etalchar{*}}25]{wang2025mega}
\textsc{Wang C., Kang D., Sun H., Qian S., Wang Z., Bao L., Zhang S.-H.}:
\newblock Mega: Hybrid mesh-gaussian head avatar for high-fidelity rendering and head editing.
\newblock In \emph{Proceedings of the Computer Vision and Pattern Recognition Conference} (2025), pp.~26274--26284.

\bibitem[WWY{\etalchar{*}}25]{wang20253d}
\textsc{Wang Y., Wang X., Yi R., Fan Y., Hu J., Zhu J., Ma L.}:
\newblock 3d gaussian head avatars with expressive dynamic appearances by compact tensorial representations.
\newblock In \emph{Proceedings of the Computer Vision and Pattern Recognition Conference} (2025), pp.~21117--21126.

\bibitem[WXL{\etalchar{*}}25]{wang2025gaussianhead}
\textsc{Wang J., Xie J.-C., Li X., Xu F., Pun C.-M., Gao H.}:
\newblock Gaussianhead: High-fidelity head avatars with learnable gaussian derivation.
\newblock \emph{IEEE Transactions on Visualization and Computer Graphics} (2025).

\bibitem[WZH{\etalchar{*}}26]{wu2026uika}
\textsc{Wu Z., Zhou B., Hu L., Liu H., Sun Y., Wang X., Cao X., Shen Y., Zhu H.}:
\newblock Uika: Fast universal head avatar from pose-free images.
\newblock \emph{arXiv preprint arXiv:2601.07603} (2026).

\bibitem[WZR{\etalchar{*}}24]{wen2024gomavatar}
\textsc{Wen J., Zhao X., Ren Z., Schwing A.~G., Wang S.}:
\newblock Gomavatar: Efficient animatable human modeling from monocular video using gaussians-on-mesh.
\newblock In \emph{Proceedings of the IEEE/CVF Conference on Computer Vision and Pattern Recognition} (2024), pp.~2059--2069.

\bibitem[XCL{\etalchar{*}}24]{xu2024gaussian}
\textsc{Xu Y., Chen B., Li Z., Zhang H., Wang L., Zheng Z., Liu Y.}:
\newblock Gaussian head avatar: Ultra high-fidelity head avatar via dynamic gaussians.
\newblock In \emph{Proceedings of the IEEE/CVF Conference on Computer Vision and Pattern Recognition} (2024), pp.~1931--1941.

\bibitem[XGGZ24]{xiang2024flashavatar}
\textsc{Xiang J., Gao X., Guo Y., Zhang J.}:
\newblock Flashavatar: High-fidelity head avatar with efficient gaussian embedding.
\newblock In \emph{Proceedings of the IEEE/CVF Conference on Computer Vision and Pattern Recognition} (2024), pp.~1802--1812.

\bibitem[XZW{\etalchar{*}}23]{xu2023latentavatar}
\textsc{Xu Y., Zhang H., Wang L., Zhao X., Huang H., Qi G., Liu Y.}:
\newblock Latentavatar: Learning latent expression code for expressive neural head avatar.
\newblock In \emph{ACM SIGGRAPH 2023 Conference Proceedings} (2023), pp.~1--10.

\bibitem[ZAB{\etalchar{*}}22]{zheng2022avatar}
\textsc{Zheng Y., Abrevaya V.~F., B{\"u}hler M.~C., Chen X., Black M.~J., Hilliges O.}:
\newblock Im avatar: Implicit morphable head avatars from videos.
\newblock In \emph{Proceedings of the IEEE/CVF Conference on Computer Vision and Pattern Recognition} (2022), pp.~13545--13555.

\bibitem[ZBT23]{zielonka2023instant}
\textsc{Zielonka W., Bolkart T., Thies J.}:
\newblock Instant volumetric head avatars.
\newblock In \emph{Proceedings of the IEEE/CVF Conference on Computer Vision and Pattern Recognition} (2023), pp.~4574--4584.

\bibitem[ZLW{\etalchar{*}}25]{zhuang2025idol}
\textsc{Zhuang Y., Lv J., Wen H., Shuai Q., Zeng A., Zhu H., Chen S., Yang Y., Cao X., Liu W.}:
\newblock Idol: Instant photorealistic 3d human creation from a single image.
\newblock In \emph{Proceedings of the IEEE/CVF Conference on Computer Vision and Pattern Recognition} (2025), pp.~26308--26319.

\bibitem[ZWL{\etalchar{*}}25]{zhang2025fate}
\textsc{Zhang J., Wu Z., Liang Z., Gong Y., Hu D., Yao Y., Cao X., Zhu H.}:
\newblock Fate: Full-head gaussian avatar with textural editing from monocular video.
\newblock In \emph{Proceedings of the IEEE/CVF Conference on Computer Vision and Pattern Recognition} (2025), pp.~5535--5545.

\bibitem[ZYW{\etalchar{*}}23]{zheng2023pointavatar}
\textsc{Zheng Y., Yifan W., Wetzstein G., Black M.~J., Hilliges O.}:
\newblock Pointavatar: Deformable point-based head avatars from videos.
\newblock In \emph{Proceedings of the IEEE/CVF Conference on Computer Vision and Pattern Recognition} (2023), pp.~21057--21067.

\bibitem[ZZWL26]{zhao2026stavatar}
\textsc{Zhao J., Zhu X., Wang Z., Lei Z.}:
\newblock Stavatar: Soft binding and temporal density control for monocular 3d head avatars reconstruction.
\newblock In \emph{Proceedings of the IEEE/CVF Conference on Computer Vision and Pattern Recognition} (2026), pp.~10996--11005.

\bibitem[ZZYH23]{zheng2023neuface}
\textsc{Zheng M., Zhang H., Yang H., Huang D.}:
\newblock Neuface: Realistic 3d neural face rendering from multi-view images.
\newblock In \emph{2023 IEEE/CVF Conference on Computer Vision and Pattern Recognition (CVPR)} (2023), IEEE, pp.~16868--16877.

\end{thebibliography}


\newpage

\end{document}